\documentclass{article}
\usepackage[preprint]{spconf}
                \copyrightnotice{\copyright\ IEEE 2019}
                \toappear{To appear in {\it Proc.\ ICASSP2019,
                    May 12-17, 2019, Brighton, UK}}
\usepackage{amsmath,bm,graphicx,multirow,amssymb,hyperref}
\usepackage[numbers,sort&compress]{natbib}
\usepackage{booktabs,url}
\usepackage{amsmath,graphicx,psfrag,pstricks,float,listings,color}
\usepackage{algorithm,algpseudocode,booktabs}
\usepackage{multirow}
\usepackage{bm}

\allowdisplaybreaks

\title{Speaker Diarisation Using 2D Self-Attentive Combination of Embeddings}
\name{{G. Sun, C. Zhang and P. C. Woodland}\thanks{Thanks to Shuai Wang and Qiujia Li for discussions and suggestions. G.~Sun is in part funded by Trinity College, Cambridge.}}
\address{Cambridge University Engineering Dept., Trumpington St., Cambridge, CB2 1PZ U.K.\\
\small{\texttt{\{gs534,cz277,pcw\}@eng.cam.ac.uk}}\\}
\begin{document}
\ninept
\maketitle


\begin{abstract}
Speaker diarisation systems often cluster audio segments
using speaker embeddings such as i-vectors and d-vectors.
Since different types of embeddings are often complementary, this paper proposes a generic framework to improve performance by combining them into a single embedding, referred to as a \textit{c-vector}. This combination uses a 2-dimensional (2D) self-attentive structure, which extends the standard self-attentive layer by averaging not only across time but also across different types of embeddings. Two types  of 2D self-attentive structure  in this paper are the simultaneous combination and the consecutive combination, adopting a single and multiple self-attentive layers respectively. The penalty term in the original self-attentive layer which is jointly minimised with the objective function to encourage diversity of annotation vectors, is also modified to obtain not only different local peaks but also the overall trends in the multiple annotation vectors. Experiments on the AMI meeting corpus show that our modified penalty term improves the d-vector relative speaker error rate (SER)   by  6\% and 21\% for d-vector systems, and a 10\% further relative SER reduction can be obtained using the c-vector from our best 2D self-attentive structure.
\end{abstract}


\section{Introduction}
\label{sec:intro}

Speaker diarisation finds ``Who spoke when" in a multi-speaker audio stream. This involves segmenting the audio into speaker-homogenous intervals followed by clustering into groups that should correspond to the same speaker.
A recent trend is for  clustering systems to first convert variable length audio segments to a fixed length vector, referred to as an embedding, and perform clustering over such vectors. More broadly, the use of embeddings has become widespread in many speech and language processing tasks.

Traditionally, i-vectors have been used as the embeddings for speech segments and are produced using factor analysis in the total variability space \cite{C1, C2,C19,C20}.  Recently, d-vectors based on outputs from an intermediate layer of deep neural networks (DNN) trained on speaker classification tasks, have shown superior performance over  i-vectors in a range of tasks \cite{C3,C4,C5,C6,C7,C9,C21,C22}. Although recent d-vector extraction systems have investigated different DNN architectures, such as deep feed-forward models \cite{C3, C5, C24, C26}, versions of recurrent neural networks (RNN)  \cite{C6, C25}, and convolutional recurrent neural networks \cite{C7}, each architecture tends to produce embeddings with different strengths and weaknesses. Therefore, it is natural to try and take advantage of the complementarity among different embeddings to achieve improved performance and robustness by combining different embedding vectors. 
For instance, a method that concatenates a d-vector and an i-vector was studied in \cite{C16}. 


In this paper, a generic framework which combines different embedding vectors via an attention mechanism is proposed.
Extending the idea of d-vector extraction using a single temporal attention model \cite{C22, C24}, the proposed attention mechanism combines outputs across both time and across embeddings from different systems and therefore has a 2D attentive structure. In particular, instead of dynamically estimating a single set of weights, a ``multi-head'' self-attentive layer \cite{C8, C10} is used for combination throughout the paper. 
The annotation vectors found in the self-attentive layer are used to find a linear combination of embeddings and are computed based on the embedding vectors themselves, and multiple annotation vectors are used to extract a more diverse combination result. The training objective function for the self-attentive layer includes a penalty term which causes the annotation vectors to be diverse. This paper modifies the penalty term so that the multiple annotation vectors produce not only distinct spiky distributions for local attention focus, but also smooth distributions that reveal overall trends. 

Two types of DNN models are studied as example d-vector systems involved in the combination, namely a feedforward  TDNN system \cite{C23} and a high order recurrent neural network (HORNN) system \cite{C12}. Two alternative structures are proposed to implement the 2D attention mechanism. First, the simultaneous attention approach where the annotation matrix produced by a single attention model is learned across all vectors extracted at each time point by each system, and second the consecutive attention approach where a separate attention across time is performed inside each system ahead of the final attention across systems. Speaker embeddings generated using this 2D attentive mechanism are named \emph{c-vectors}. Experimental results on speaker clustering show that our modified penalty term improves d-vector extraction by a clear margin, and c-vectors with both 2D attentive structures outperform the individual d-vector systems with the consecutive structure gives the lowest error rate.

The remainder of this paper is as follows: Section \hyperref[sec:pipeline]{2} reviews the self-attentive structure. Two types of 2D attentive combination along with the modified penalty term are introduced  in Sec. \hyperref[sec:archi]{3}. Experimental setup and results are in Secs. \hyperref[sec:exp]{4} and  \hyperref[sec:res]{5} followed by conclusions.

\section{Self-attentive Structure}
\label{sec:pipeline}

An important step of d-vector generation is the combination of embedding vectors $\mathbf{h}(t)$ extracted at each time $t$ in a window of several seconds. As these embedding vectors may have different level of speaker-discriminative ability, they should be combined with different weights. Therefore, a self-attentive layer is introduced to achieve a dynamic linear combination, where an annotation matrix $\mathbf{A}$, computed from the input vectors, provides the combination weights. Each column of the annotation matrix is an annotation vector which gives a set of scaling factors that weights the importance of each input before summing them. Specifically, if $T$ input vectors within a window forms a $T\times n$ matrix $\mathbf{H}  =[\mathbf{h}(1), \mathbf{h}(2), \dots, \mathbf{h}(T)]^T$ where $n$ is the dimension of each vector, the annotation matrix can be calculated using Eq.~(\ref{eq:eq1}) and applied to the inputs as in Eq.~(\ref{eq:eq2})

\begin{equation}
\mathbf{A} = \text{Softmax}(\tanh(\mathbf{H}\mathbf{W_1})\mathbf{W_2}),
\label{eq:eq1}
\end{equation}
\begin{equation}
\mathbf{E} = \mathbf{A}^T\mathbf{H},
\label{eq:eq2}
\end{equation}
where $\mathbf{E}$ ($h\times n$) is the output and $\mathbf{A}$ ($T\times h$) is the  $h$-head annotation matrix. $\mathbf{A}$ is generated by passing the input matrix through two fully-connected layers with weight matrices $\mathbf{W_1}$ and $\mathbf{W_2}$ respectively, and the Softmax is performed column-wise to ensure each annotation vector sums to one. When a multi-head self-attentive layer is used (i.e. $h>1$), to encourage different heads to extract dissimilar information, 
a penalty term in Eq.~(\ref{eq:eq3}) is added to the cross-entropy loss function during training.
 \begin{equation}
P = \mu{||\mathbf{A}^T\mathbf{A} - \mathbf{I}||}^2_F = \mu\Big{(}\sum_i^h(\mathbf{a_i}^T\mathbf{a_i}-1)^2 + \sum_{\substack{i,j, i\neq j}}^h(\mathbf{a_i}^T\mathbf{a_j})^2\Big{)},
\label{eq:eq3}
\end{equation}
 where $\mathbf{a_i}$ is the annotation vector, $\mathbf{I}$ is the identity matrix and $||\cdot||_F$ denotes the Frobenius norm. The degree of influence of this term is adjusted by $\mu$. As all terms in Eq.~(\ref{eq:eq3}) are non-negative, minimising the cross terms, $(\mathbf{a_i}^T\mathbf{a_j})^2$, encourages the annotation vectors to be orthogonal, while minimising the diagonal terms $(\mathbf{a_i}^T\mathbf{a_i}-1)^2$ encourages the annotation vectors to have fewer non-zero terms, ideally being one-hot vectors. Therefore, under the effect of the penalty term, the annotation vectors will put large weights on different but very few inputs. However, the trend of each annotation vector, whether to be spiky or smooth, can also be controlled, see. Sec.~\hyperref[ssec:penalty]{3.3}. In the rest of the paper, the output from the self-attentive structure in a single system is denoted using $\mathbf{E}$,  as in \cite{C10}, while that incorporating model combination is denoted using $\mathbf{C}$, for c-vectors. For clarity, the output of $h$-head self-attentive layer is expressed as:
\begin{align}
\mathbf{E} = [\mathbf{e}_1, \mathbf{e}_2 ... \mathbf{e}_h] = \text{SelfAtten}\big(\mathbf{h}(1), \mathbf{h}(2), \dots, \mathbf{h}(T)\big).
\label{eq:eq4}
\end{align}
\section{2D Self-attentive Topologies}
\label{sec:archi}
In this section, the two kinds of 2D self-attentive structures are introduced, and the modification to the penalty term is also explained.
\subsection{Simultaneous Combination Architecture}
\label{ssec:simult}

\begin{figure}[th]
    \centering
    \includegraphics[width=\linewidth]{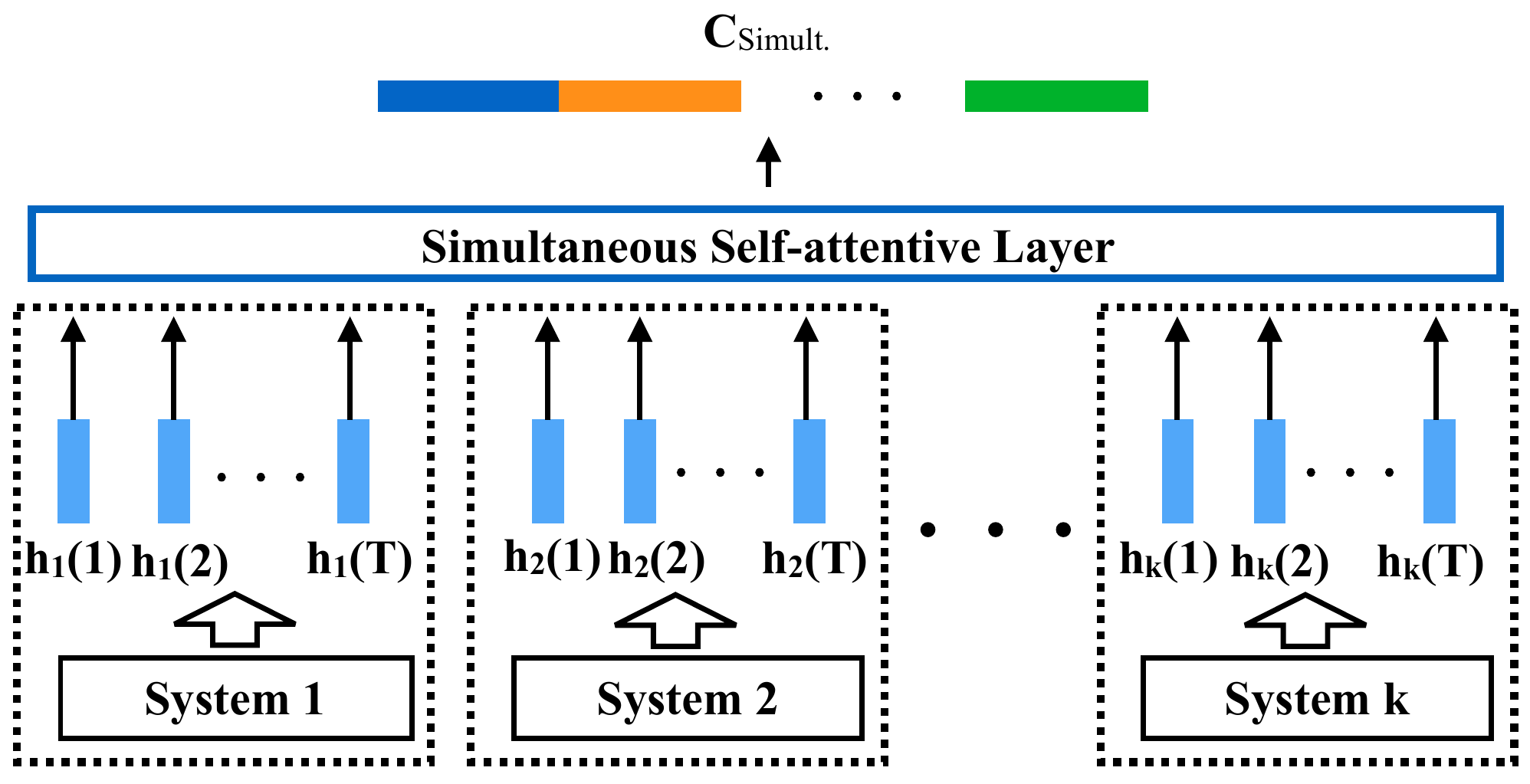}
    \vspace{-2em}
    \caption{c-vector with the simultaneous combination architecture.}
    \vspace{-1em}
    \label{fig:simul_atten}
\end{figure}
One natural way of performing the 2D attention is to add up all the embeddings extracted from each frame by each network simultaneously using one annotation matrix. This is achieved by extending the row dimension of matrix $\mathbf{A}$ in Eq.~(\ref{eq:eq1}) from $T$ to $k \times T$ where $k$ is the number of systems to be combined, as shown in Eq.~(\ref{eq:eq5}).
\begin{equation}
\mathbf{C}_\text{Simult.} = \text{SelfAtten}\big(\mathbf{h}_1(1), \dots, \mathbf{h}_i(t), \dots, \mathbf{h}_k(T)\big).
\label{eq:eq5}
\end{equation}
The c-vector of this combination is the concatenation of the embeddings each generated using one annotation vector, as shown in Fig.~\ref{fig:simul_atten}. This structure is not only able to reflect the importance of each frame in terms of the ability to distinguish speakers, but also able to weight the two network outputs frame by frame.

\subsection{Consecutive Combination Architecture}
\label{ssec:consec}
\begin{figure}[h]
    \centering
    \includegraphics[width=\linewidth]{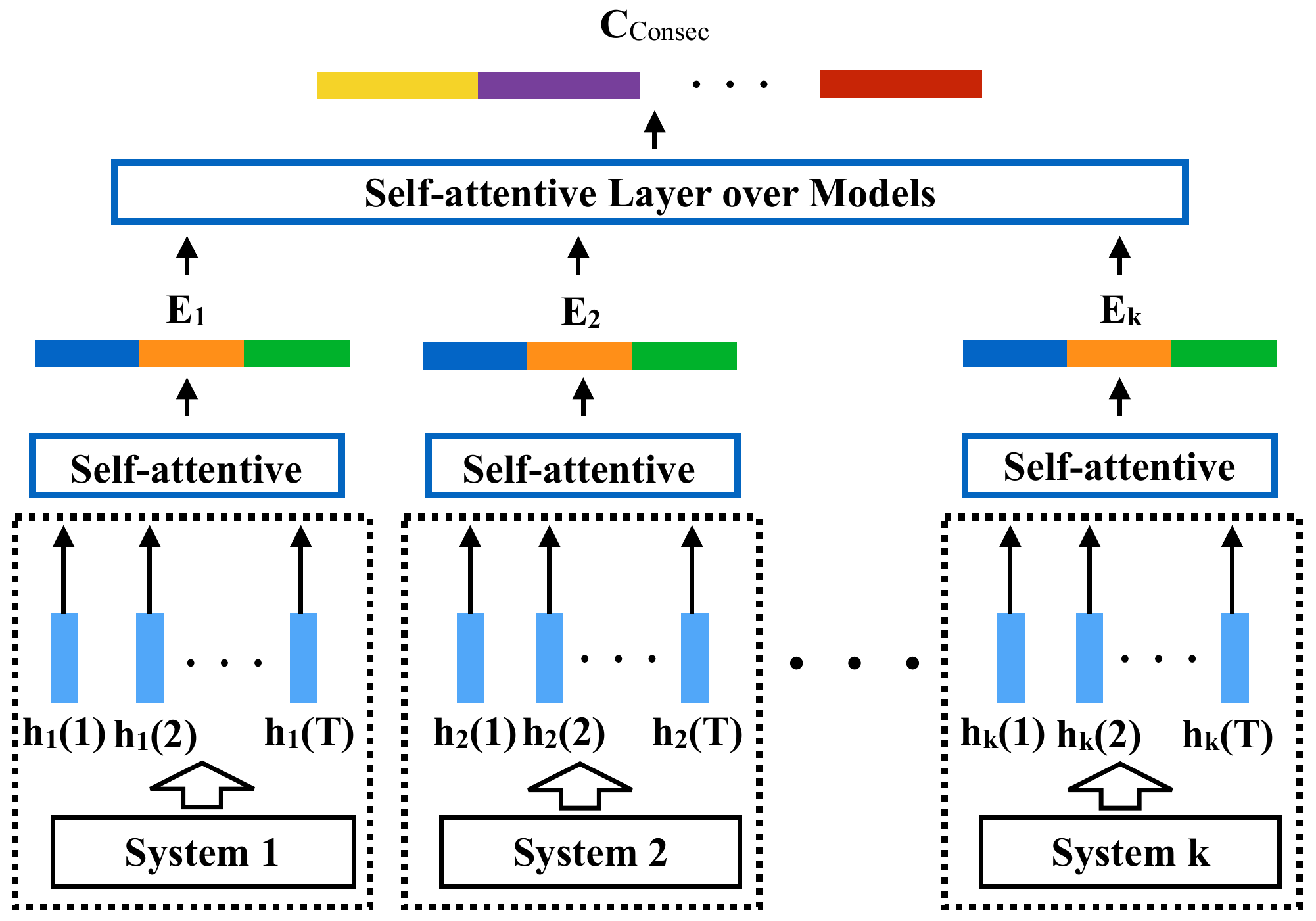}
    \vspace{-2em}
    \caption{c-vector with the consecutive combination architecture.}
    \label{fig:consec_atten}
    \vspace{-1em}
\end{figure}

An alternative proposed uses consecutive combination where self-attentive combination is first performed across time for each system with separate annotation matrices, and another self-attentive layer is applied across all the systems thereafter, as illustrated in Fig.~\ref{fig:consec_atten}. As the self-attentive layers for each system can be designed differently, this combination retains more individuality of each system while introducing more flexibility.

Particularly, it is  interesting to investigate the following two types of second stage combination. First, model combination could be performed on the multi-head output where all heads in the d-vector share the same annotation vector, as shown in Eq.~(\ref{eq:eq6}).
\begin{equation}
\mathbf{C}_{\text{Consec1}} = \text{SelfAtten}\big(\mathbf{E}_1, \dots, \mathbf{E}_i, \dots, \mathbf{E}_k\big),
\label{eq:eq6}
\end{equation}
where $\mathbf{E}_i$ is the multi-head output generated from each individual system. Secondly, it could also be performed at the head level where different heads from the same system can be assigned different weights, as shown in Eq.~(\ref{eq:eq7}). This relaxes the constraint on the number of heads for each system which has to be equal in the previous combination method. 
\begin{equation}
\mathbf{C}_{\text{Consec2}} = \text{SelfAtten}\big(\mathbf{e}_{11}, \dots,\mathbf{e}_{1h}, \dots,\mathbf{e}_{k1}, \dots,\mathbf{e}_{kh}\big),
\label{eq:eq7}
\end{equation}
As the aspects of speaker characteristic information encapsulated in the d-vectors for different systems may be ordered differently, a direct weighted average of the output vectors may be inappropriate. Therefore, a fully-connected (FC) layer that transforms the output of each system is introduced before the model combination, as shown in Eq.~(\ref{eq:eq8}).
\begin{equation}
\mathbf{E}^*_i = \text{ReLU}(\mathbf{W}^T \mathbf{E}_i).
\label{eq:eq8}
\end{equation}
 Further extending the method above, instead of using self-attention for model combination, embeddings from different systems are concatenated and passed through an FC layer for transformation and combination together, as shown in Eq.~(\ref{eq:eq9}). 
\begin{equation}
\mathbf{C}_{\text{ConsecFC}} = \text{ReLU}\big(\mathbf{W}^T[\mathbf{E}_1, \dots, \mathbf{E}_i, \dots, \mathbf{E}_k]\big),
\label{eq:eq9}
\end{equation}
A  similar approach was proposed in \cite{C16} for speaker verification tasks  where the i-vector is fused with an RNN output by direct concatenation. Nevertheless, the proposed c-vector method allows joint training of the complete model, and by including the FC layer, the order of elements in e.g. an i-vector could also be altered.

\subsection{Penalty Term Modification}
\label{ssec:penalty}

The penalty term for multi-head self-attention in Eq.~(\ref{eq:eq3}) was originally designed for sentence embeddings \cite{C8} to focus on as few words as possible while encouraging different annotation vectors to be estimated. Such a setting is not necessarily transferable to our task, since the minimum value of the penalty term can be reached only when all the annotation vectors become different one-hot vectors. Therefore, propose the following modified penalty term is proposed:

\begin{equation}
P = \mu{||\mathbf{A}^T\mathbf{A} - \bm{\Lambda}||}^2_F,
\label{eq:eq10}
\end{equation}
where $\bm{\Lambda}$ is a diagonal matrix replacing $\mathbf{I}$ in Eq.~(\ref{eq:eq3}). The diagonal values $\lambda_i = \bm{\Lambda}_{ii}$ control the smoothness of the annotation vectors. We term annotation vectors that only focus on a few input vectors ``spiky",  while ``smooth" annotation vectors reflect the general trends of importance. For a single annotation vector, $\mathbf{a}$, the penalty term becomes a quadratic function of $\lambda$. For the annotation vector $\mathbf{a}$ taking one-hot form and more evenly distributed forms, the variation of the penalty term P against $\lambda$ is plotted below. 
\begin{figure}[h]
    \centering
    \includegraphics[width=0.9\linewidth]{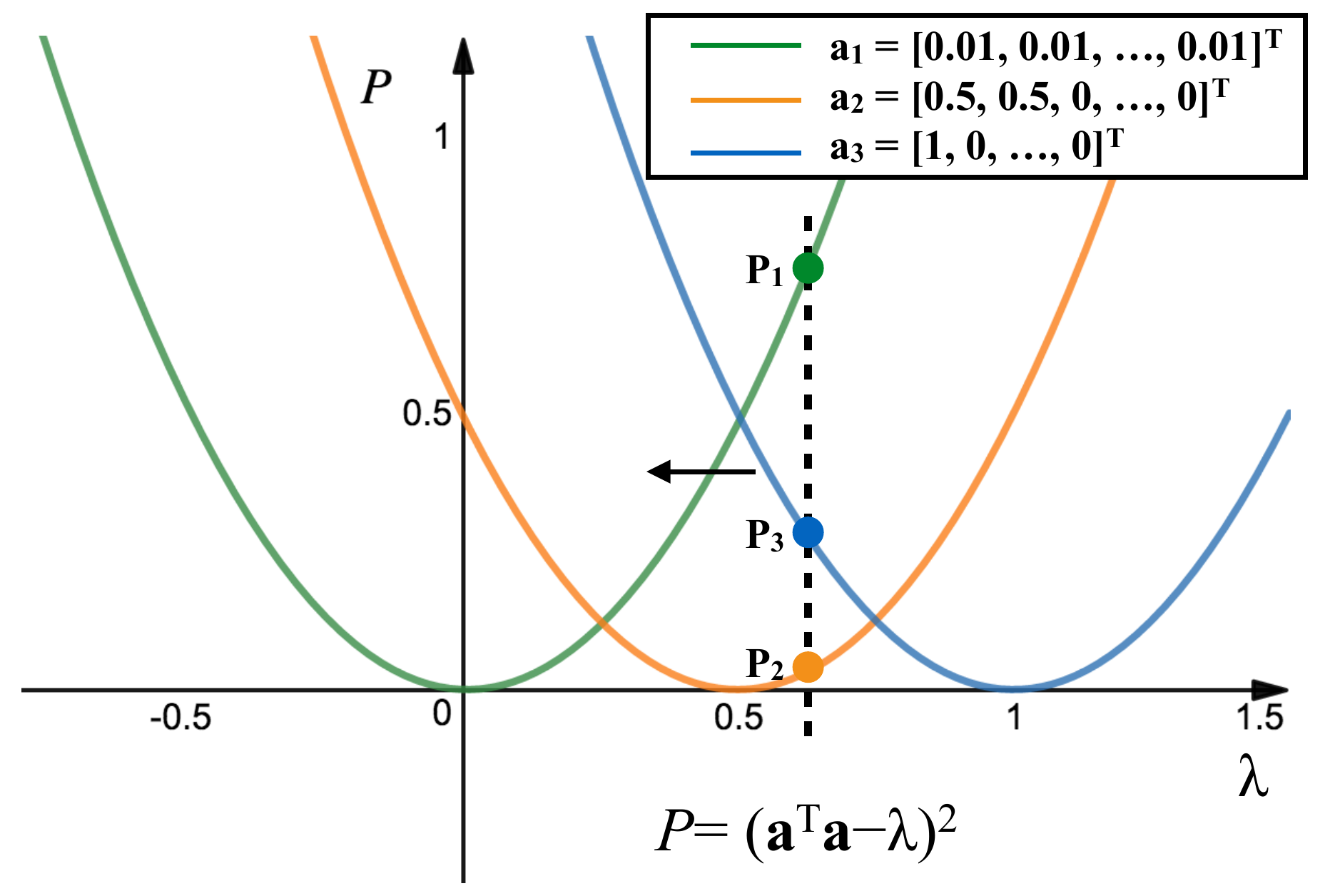}
    \vspace{-1em}
    \caption{Penalty term $P$ with different $\lambda$ can have its minimum value shifted, to generate ``spiky'' or ``smooth'' annotation vectors.}
    \label{fig:penalty}
\end{figure}

As the dashed line moves toward the left, which represents decreasing $\lambda$, the lowest point changes from $\mathbf{P_3}$ to $\mathbf{P_2}$ to $\mathbf{P_1}$, and hence the annotation vector that gives the minimum value of the penalty term shifts from $\mathbf{a}_3$ to $\mathbf{a}_2$, and eventually reaches the evenly distributed $\mathbf{a}_1$. Therefore, by varying the value of $\lambda$ between $1/T$ which is the $l2$-norm of uniform vector $\mathbf{a}_1$, and 1 which is the $l2$-norm of one-hot vector $\mathbf{a}_3$, the smoothness of the weight can be controlled. The multi-head system can use this modified penalty term to give a couple of smooth annotation vectors while keeping the rest spiky as before, and these settings will also vary between systems according to their characteristics.

\section{Experimental Setup}
\label{sec:exp}

\subsection{Data Preparation}

All of the models were implemented using an extended version of HTK \cite{C27}, and trained and tested on the AMI corpus \cite{C18} which contains group meetings recorded at four different sites. 
The full training  set which contains 135 meetings with 149 speakers was used which is further split into 90\% for model training and 10\% cross validation set for hyper-parameter tuning. 
For evaluation, instead of using the full dev and eval sets, we use the meetings recorded at IDIAP, Edinburgh and Brno which are the sets frequently used for evaluation of speaker diarisation \cite{C7,C21}, and which are 
more consistent with our observations on other datasets. The partition of the dataset is shown in {Table \ref{table:set}}. 

\begin{table}[H]
\centering
\begin{tabular}{ccl}
\toprule
&Meetings & Speakers\\
\midrule
\centering
Train & 135 & 149\\
\centering
Dev   & 14 & 17 (4 seen in Train)\\
\centering
Eval  & 12 & 12 (0 seen in Train)\\
\bottomrule
\end{tabular}
\caption{Details of the AMI data set based on official speech recognition partition with the TNO meetings excluded from Dev and Eval.}
\vspace{-1.0em}
\label{table:set}
\end{table}
During both training and testing, the system input is 40-d log-mel filter bank features (25 ms frame size, 10 ms frame increment) extracted from Multiple Distance Microphone (MDM) data after beam-forming using \textit{BeamformIt} \cite{C28}. 
\begin{table*}[t]
  \centering
\begin{tabular}{p{3.0cm}p{1.5cm}p{1.5cm}p{1.5cm}p{1.5cm}p{1.5cm}p{1.5cm}p{1.5cm}p{1.5cm}}
\toprule
\centering
 & DiarTK & d-vector TDNN & d-vector HORNN & c-vector {~}Simult. & c-vector Consec. 1 & c-vector Consec. 2 & c-vector Consec. FC \\
\midrule
\centering
 \#Params. & N/A & 1.76M & 0.29M & 2.03M & 2.46M & 2.07M & 2.87M\\
\centering
Dev   & 23.62\% & 13.40\% & 13.40\% &12.73\% & 13.18\% & \textbf{12.22\%} & 12.75\%\\
\centering
Eval  & 23.31\% & 14.75\% & 15.97\% &16.28\% & 13.53\% & \textbf{12.99\%} & 15.00\%\\
\toprule
\end{tabular}
\caption{Speaker Error Rate (SER) on Dev and Eval sets. ``\#Params.'' is the number of parameters in million (M) used in d-vector/c-vector extraction. Simult. and Consec. refer to the simultaneous and consecutive combination architectures. }
\label{table:reduce}
\end{table*}

\subsection{Model Specification}
The two DNN systems used as an example for combination in this paper are the TDNN and HORNN. The TDNN structure resembles the one used in the x-vector extraction system \cite{C9}, except the statistical pooling layer is replaced with self-attentive layer as described in \cite{C10}. The HORNN here uses ReLU activation functions, and adds connections from both the previous hidden state and the state with 4 time steps ahead to the current RNN input. This provides a more direct access to the long-term memory to prevent vanishing gradient problem with much less parameters than the LSTM structure.

To extract window level d-vectors, a 2-second sliding window was applied with a 1-second overlap between adjacent windows.  A two-layer HORNN was used with a state output dimension of 256 and a projection dimension of 128. For the TDNN, the original 512-dimensional system in \cite{C10} was used. In order to have similarly performing systems, the HORNN uses fewer parameters than the TDNN, as the former uses parameters in a more efficient way. Then, both network outputs are reduced to 128-d vectors using the fifth layer of the TDNN and an additional fully-connected layer for HORNN before feeding into the self-attentive layer. The simultaneous combination learns a set of 5 weight annotation vectors. The consecutive combination have 5 heads from each system, and the second combination stage uses a single head and 5 heads for the first and second types of attentive combinations respectively.

After the combination stages, we use a bottleneck layer to map the multi-head c-vector output down to a 128-dimensional representation space, which is then used as the c-vector for clustering. Two individual networks are initialised with frame-level pre-training, and then jointly trained in the combination networks. Furthermore, instead of using normal softmax at the output layer, we adopt the "Asoftmax" function \cite{C15} in the $m = 1$ case to provide better discrimination in the angular aspect. This further helps the clustering process as it constructs the affinity matrix using cosine distances.

\subsection{Diarisation Pipeline}
 As the main focus of the paper is on the use of new speaker embeddings in clustering, similar to e.g.  \cite{C21,C29,C30}, the experiments reported here use the AMI manual segments and report only the speaker error (there is no missed or false alarm speech). As in training, a 2-second sliding window with 1-second overlap is applied to the segments, and c-vectors extracted by forward propagation to the bottleneck layer. These window level embeddings are then clustered using the spectral clustering methods proposed in \cite{C6}. The threshold value used in the affinity matrix pre-processing stage is tuned for each system separately on the dev set, and applied to the eval set. Scoring uses the setup from NIST-RT evaluations with a 0.25 second collar.

\subsection{Baseline Systems}
The first baseline used DiarTK  \cite{C14} to perform bottom-up agglomerative clustering based on the information bottleneck principle \cite{C13,C17}. It used 19-d MFCCs and the maximum window length to be 2 seconds in DiarTK.
The values of $\beta$ and NMI threshold were set to be 10 and 0.3 respectively. Another baseline uses the statistical pooling layer \cite{C9} in the two example DNNs which calculates the mean and standard deviation across the frames instead of using the self-attentive layer.

\section{Results}
\label{sec:res}

The reductions in SER obtained for TDNN and HORNN by using the modified penalty term are shown in {Table \ref{table:weightres}}. 

\begin{table}[h]
\centering
\begin{tabular}{p{1.2cm}p{1.2cm}p{1.2cm}p{1.2cm}p{1.4cm}}
\toprule
\centering
 & Dataset & Mean+std. deviation& Attention (original) & Attention (modified)\\
\midrule
\centering
 \multirow{2}{*}{HORNN} & Dev   & 21.00\% & 16.72\% & \textbf{13.40\%}\\
\centering
& Eval  & 23.70\% & 20.55\% & \textbf{15.97\%}\\
\midrule
\centering
 \multirow{2}{*}{TDNN} & Dev   & 17.46\% & 15.02\% & \textbf{13.40\%}\\
\centering
& Eval  & 19.22\% & 14.95\% & \textbf{14.75\%}\\
\bottomrule
\end{tabular}
\caption{d-vector system SERs with different embedding extraction schemes. ``Attention'' columns are self-attentive layers with the original (5 spiky) and our modified (3 spiky$+$2 smooth) penalty terms.}
\label{table:weightres}
\end{table}

\begin{figure}[h]
\begin{minipage}[b]{0.9\linewidth}
    \centering
    \includegraphics[width=0.9\linewidth]{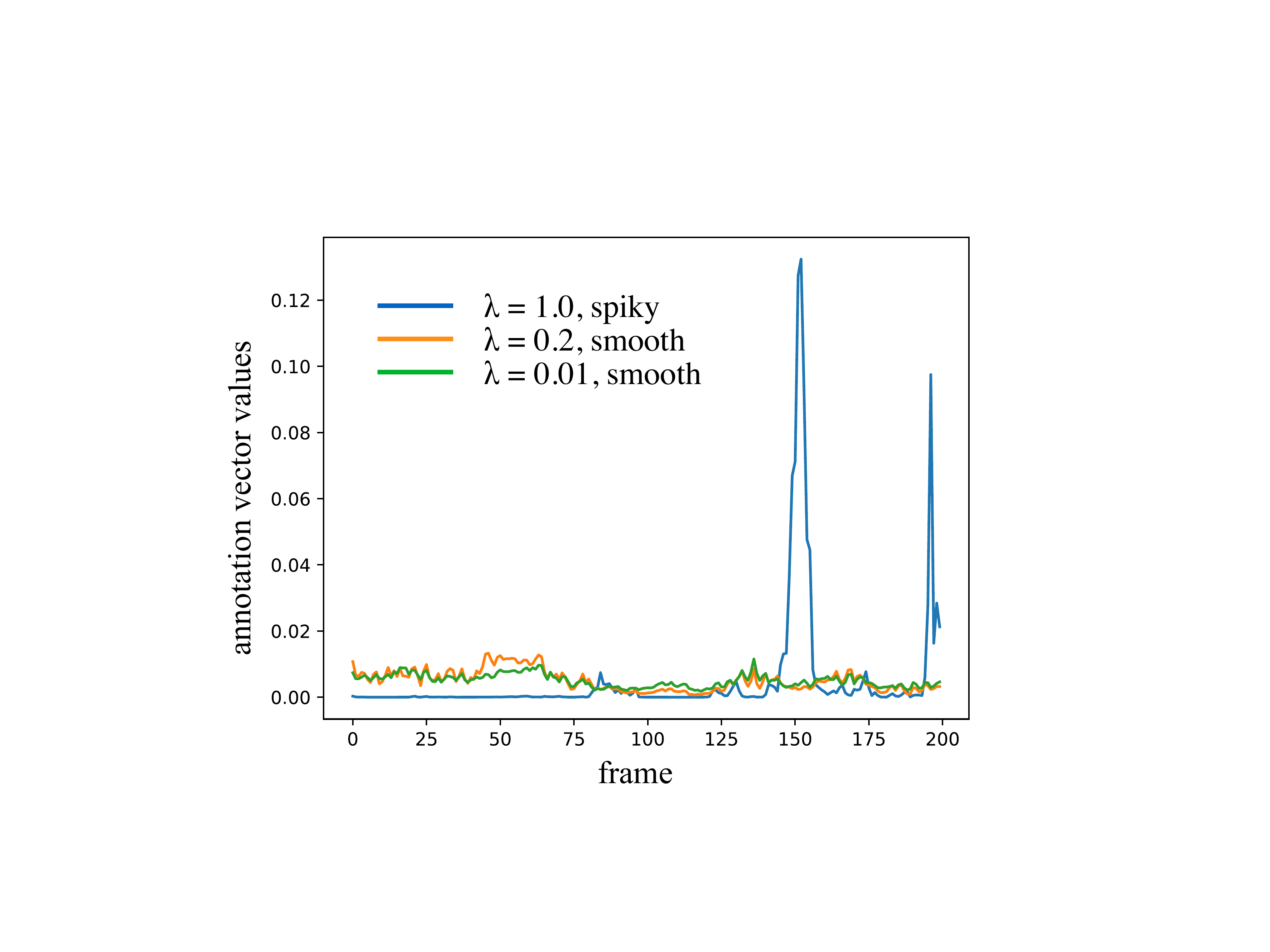}
    \vspace{-1.0em}
    \caption{An example of the annotation vectors obtained with the modified penalty term (1 spiky$+$2 smooth shown).}
    \label{fig:weightvecs}
    \end{minipage}
    \vspace{-1.0em}
\end{figure}

Compared to the d-vector using mean and standard deviation, there were reductions in SER using the self-attentive layer with the original penalty term. The modified penalty term further gives a relative reduction in SER  for the TDNN by 6\% and for the HORNN system 21\%. The effect of changing $\lambda$ in the penalty term can be seen in Fig. \ref{fig:weightvecs}, where each curve represents one annotation vector across 200 frames in a selected window corresponding to a  specific $\lambda$ value. The curve with $\lambda = 1.0$ provides two spikes at the 150-th and 190-th frames respectively, while the curve with $\lambda = 0.2$ reflects a general trend of which regions of frames are more important. 

The results of using different 2D attentive combinations are shown in  {Table \ref{table:reduce}} above\footnote[1]{The models were also tested on the full dev and eval sets. Improvements were found using the 2D attentive combinations and FC layer combination.}. Even though the HORNN system has far fewer parameters than the TDNN, they provide similar performance on the dev set where optimised system-specific threshold values were used. 
{Table~\ref{table:reduce}} shows that all combinations achieve improvements on the dev set where the clustering threshold is optimised. The performance on the eval set is rather more variable, but both types of consecutive combination show their superiority over the individual d-vector systems. In particular, the second method of consecutive model combination achieves a consistent relative reduction in SER of 9\% and 12\% on dev and eval set respectively, and 10\% overall relative reduction, which provides the best performance. This represents a 46\% reduction in SER over the DiarTK baseline.

\section{Conclusions}
\label{sec:conc}

In this paper, a novel embedding extraction approach for diarisation using a 2D self-attentive structure has been proposed. Both simultaneous combination and consecutive combination approaches were analysed. Furthermore, a modified penalty term was also introduced which provided more diversity to the multi-head weight vector in the self-attentive layer. Taking the TDNN and HORNN as an example of two complementary systems, the proposed models were evaluated in experimented using the AMI corpus. Experimental results showed a relative reduction in diarisation speaker error rate of 21\% for a HORNN model and 6\% for a TDNN  model by including the modification to the penalty term. Furthermore, a further reduction in SER of 10\% was obtained using the 2D consecutive combination method.


\vfill\pagebreak

\renewcommand{\bibsection}{}
\section{REFERENCES}

\end{document}